\documentclass[10pt,conference]{IEEEtran}
\usepackage[utf8]{inputenc}
\usepackage{balance}
\usepackage{cite}
\usepackage{amsthm}
\usepackage{multirow}
\usepackage{algorithm}
\usepackage{algorithmic}
\usepackage{float}
\usepackage[table,xcdraw]{xcolor}
\usepackage{xspace}
\usepackage{graphicx}
\usepackage{todonotes}
\usepackage{amsfonts}
\usepackage{amsmath,amssymb,cuted}
\usepackage{booktabs}
\newcolumntype{N}{>{\centering\arraybackslash}m{.5in}}
\newcolumntype{G}{>{\centering\arraybackslash}m{2in}}
\def\BibTeX{{\rm B\kern-.05em{\sc i\kern-.025em b}\kern-.08em
    T\kern-.1667em\lower.7ex\hbox{E}\kern-.125emX}}
\usepackage{acro}
\usepackage{enumitem}
\usepackage{soul}
\usepackage{mathtools,lipsum, nccmath,cuted,amsmath}
\usepackage[capitalise]{cleveref}
\usepackage{siunitx}
\usepackage{textcomp}

\usepackage{comment}
\usepackage{makecell}
\usepackage{gensymb}
\usepackage[normalem]{ulem}
\usepackage{subcaption}
\usepackage{setspace}

\newlength{\tempdima}
\renewcommand{\thesubfigure}{\alph{subfigure}}

\newcommand{\mycaption}[1]
{\refstepcounter{subfigure}\textbf{(\thesubfigure) }{\ignorespaces #1}}
\newcommand{\rowname}[1]
{\rotatebox{90}{\makebox[\tempdima][c]{#1}}}

\usepackage{array}

\usepackage[font=footnotesize]{caption}
\usepackage{pgfplotstable}
\usepackage{soul}
\usepackage{xcolor}
\usepackage{multirow}

\newcolumntype{M}[1]{>{\centering\arraybackslash}m{#1}}
\author{
    \IEEEauthorblockN{{Matthew Burruss\IEEEauthorrefmark{1}\textsuperscript{\textsection}, Shreyas Ramakrishna\IEEEauthorrefmark{2}, and Abhishek Dubey\IEEEauthorrefmark{2}}
    \IEEEauthorblockA{\IEEEauthorrefmark{1} Microsoft Corporation, \IEEEauthorrefmark{2} Vanderbilt University}
}}

\DeclareAcronym{cps}{
  short = CPS,
  long = Cyber-Physical Systems,
}

\DeclareAcronym{gtsb}{
  short = GTSB,
  long = German Traffic Sign Benchmark,
}

\DeclareAcronym{rbf}{
  short = RBF,
  long = Radial Basis Function,
}

\DeclareAcronym{cnn}{
  short = CNN,
  long = Convolutional Neural Network,
}

\DeclareAcronym{dnn}{
  short = DNN,
  long = Deep Neural Network,
}

\DeclareAcronym{e2e}{
  short = e2e,
  long = end-to-end,
}

\DeclareAcronym{lec}{
  short = LEC,
  long = Learning Enabled Component,
}

\DeclareAcronym{ood}{
  short = OOD,
  long = Out-of-Distribution
}

\DeclareAcronym{am}{
  short = AM,
  long = Assurance Monitor
}

\DeclareAcronym{vae}{
  short = VAE,
  long = Variational Autoencoder
}

\begin{document}


\title{Deep-RBF Networks for Anomaly Detection in Automotive Cyber-Physical Systems}


\maketitle
\begingroup\renewcommand\thefootnote{\textsection}
\footnotetext{Work performed during Master Thesis at Vanderbilt University.}
\endgroup

\pagestyle{plain}
\begin{abstract}
Deep Neural Networks (DNNs) are widely used in automotive Cyber-Physical Systems (CPS) to implement autonomy related tasks. However, these networks have exhibited erroneous predictions to anomalous inputs that manifest either due to Out-of-Distribution (OOD) data or adversarial attacks. To detect these anomalies, a separate DNN called assurance monitor is used in parallel to the controller DNN, increasing the resource burden and latency. We hypothesize that a single network that can perform controller predictions and anomaly detection is necessary to reduce the resource requirements. Deep-Radial Basis Function (RBF) networks provide a rejection class alongside the class predictions, which can be used for anomaly detection. However, the use of RBF activation functions limits the applicability of these networks to only classification tasks. In this paper, we discuss the steps involved in detecting anomalies in CPS regression and classification tasks. Further, we design deep-RBF networks using popular DNNs such as NVIDIA DAVE-II and ResNet20 and then use the resulting rejection class for detecting physical and data poison adversarial attacks. We show that the deep-RBF network can effectively detect these attacks with limited resource requirements.

\end{abstract}


\section{Introduction}
\label{sec:introduction}
\begin{figure}[t]
 \includegraphics[width=0.9\columnwidth]{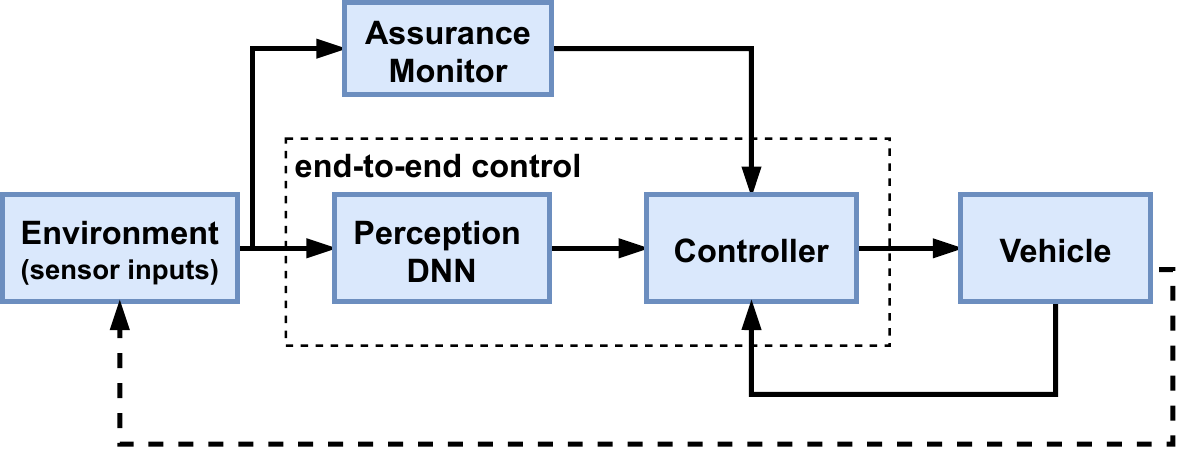}
 \centering
 \caption{end-to-end automotive CPS architecture with an assurance monitor. The monitor receives the same sensory input as the controller \ac{dnn}, and its detection results are often used in control decision making.}
 \vspace{-0.2in}
 \label{fig:cps}
\end{figure}

\textbf{Emerging Trend}: \acp{dnn} are widely used in automotive \acp{cps} to implement autonomy related tasks. One way to use these networks in autonomous driving is in an \ac{e2e} fashion, where the network takes in sensory inputs to predict control actions (e.g. steer) as shown in \cref{fig:cps}. A well-known example of an \ac{e2e} network is NVIDIA's DAVE-II \ac{cnn}~\cite{bojarski2016end} which steered a car autonomously. Despite being widely used, \acp{dnn} demonstrates susceptibility to anomalies that manifest as \ac{ood} data or adversarial attacks. To detect these anomalies, a separate \ac{dnn} called \ac{am} is often trained and used in parallel to the \ac{dnn} controller as shown in \cref{fig:cps}. These monitors identify if the operational test inputs to the \ac{dnn} belongs to the training distribution.

\textbf{State-of-the-art and Challenges}: Recently, there has been a growing interest in using Generative models like Generative Adversarial Network (GAN) and \ac{vae} as assurance monitors for detecting \ac{dnn} related anomalies. Although these monitors have shown robust performance for anomaly detection in \acp{cps}~\cite{cai2020real,sundar2020out}, they introduce additional resource and time overhead as found in our previous work~\cite{hartsell2021resonate}. We hypothesize that a single \ac{dnn} to perform both controller predictions and anomaly detection is necessary for \acp{cps} which usually have limited resources and short inference times. In this direction, Amini, Alexander \emph{et al.}~\cite{amini2018variational} have used a single \ac{vae} to perform continuous steering predictions and anomaly detection. However, generative models are dependent on several hyperparameters, and training them is challenging (e.g. mode collapse problem of GAN). 

Deep-\ac{rbf} networks~\cite{goodfellow2014explaining} provide a rejection class alongside the class predictions, which can be utilized for anomaly detection. These networks are conventional \acp{dnn} with an output \ac{rbf} layer and do not have additional hyperparameters to tune. Recently, the authors in~\cite{zadeh2018deep} and~\cite{crecchi2020fader} have used the rejection capability of these networks to detect adversaries in toy classification datasets such as MNIST and CIFAR10. However, to the best of our knowledge, it has yet to be shown whether these networks are capable of detecting anomalies in the \ac{cps} domain. Additionally, these networks are mostly designed for classification tasks, limiting their utility in regression tasks. 

\textbf{Our Contributions}: In this work, we design a single deep-\ac{rbf} network for predicting control actions (e.g. steering) and detecting anomalies (especially adversarial attacks) in \ac{cps} regression tasks. We hypothesize that the non-linearity introduced by the \ac{rbf} layer decreases the network's susceptibility to anomalies while increasing its confidence in recognizing in-distribution data and consequently rejecting \ac{ood} data. However, as these networks are limited to classification tasks, we discuss the steps required for transforming a \ac{cps} regression task (continuous steering) to a classification task (discrete steering classes). We then integrate \acp{rbf} to the output layer of well-known \acp{dnn} such as NVIDIA's DAVE-II and ResNet20 and train the resulting deep-\ac{rbf} network. We then use the rejection class of the trained network to detect adversarial attacks, including physical attacks and data poison attacks. We craft the physical attack on a hardware \ac{cps} testbed called DeepNNCar ~\cite{ramakrishna2019augmenting} and the data poison attack on the real-world traffic sign dataset called \ac{gtsb}. We show that the deep-\ac{rbf} network can detect these attacks with limited resource requirements.

\section{Related Work}
\label{sec:rw}

\acp{dnn} are widely used as \acl{am}s for detecting \ac{cps} related anomalies that manifest either due to \ac{ood} data or adversarial attacks. Generative models such as Generative Adversarial Networks (GANs), and variants of Autoencoders (e.g. \ac{vae}) have been recently used to detect \ac{dnn} related anomalies~\cite{sundar2020out,cai2020real}. Despite robust detection capabilities, these models require an independent \ac{dnn} trained in parallel to the controller \ac{dnn}, which adds excessive resource burden and latency as observed in our previous work~\cite{hartsell2021resonate}. 

To address this, Amini, Alexander \emph{et al.}~\cite{amini2018variational} have used a single \ac{vae} network for both controller predictions and anomaly detection. Though this is significant for reducing the resource burden, the encoder-decoder architecture of the \ac{vae} is still resource expensive. Additionally, the detection efficiency of the \ac{vae} depends on hyperparameters (e.g. size of the latent space) which have no default values. Deep-\ac{rbf} networks are another variant of \acp{am} that have shown robustness in detecting anomalies, especially adversarial attacks~\cite{goodfellow2014explaining}. Zadeh \emph{et al.}~\cite{zadeh2018deep} has shown a single deep-\ac{rbf} network is capable of predicting both class predictions and anomalies without the need for tuning complex hyperparameters. These networks have a \ac{dnn} structure with \ac{rbf} attached to the output layer. The classification capability of the \ac{rbf} along with a threshold is used as a rejection class to detect adversarial attacks on trivial classification datasets such as MNIST~\cite{zadeh2018deep}, and CIFAR10~\cite{crecchi2020fader}. In this work, we transform a \ac{cps} regression task to classification and evaluate the deep-\ac{rbf} network's anomaly detection capability for runtime \ac{cps} applications.

\begin{figure*}[t]
 \includegraphics[width=0.85\textwidth]{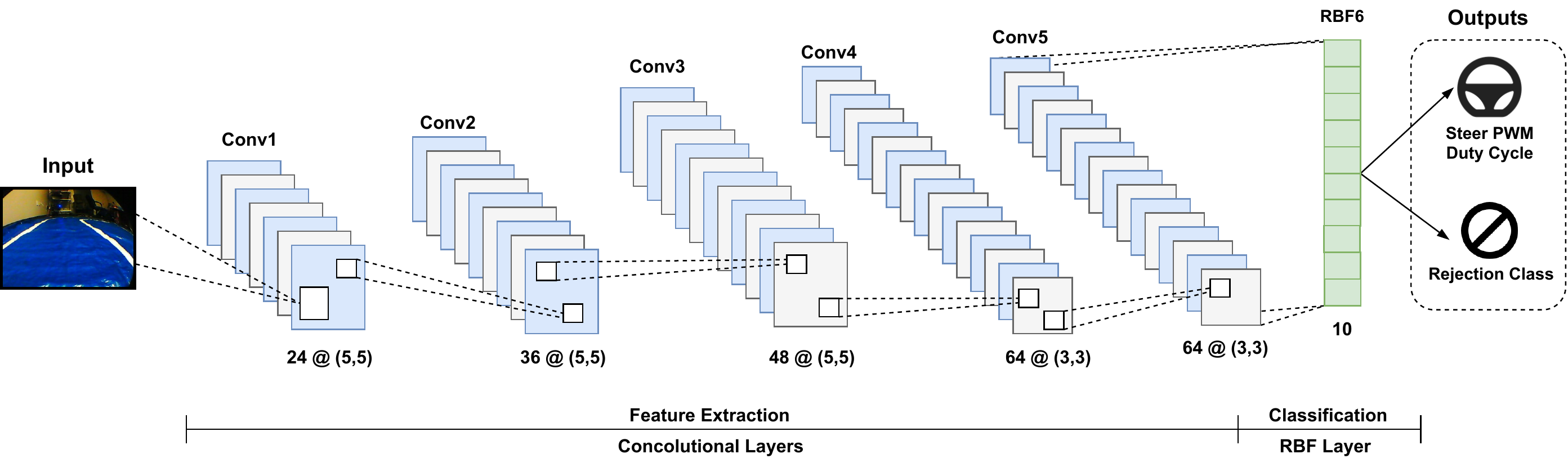}
 \centering
 \caption{Deep-\ac{rbf} network for \ac{e2e} control of the DeepNNCar example. The convolutional layers extract the image features, which are sent to the \ac{rbf} layer to perform classification. The outputs are the class predictions (discrete steering values) and a rejection class to detect anomalies.}
 \label{fig:DAVEIIArchitecture}
 \vspace{-0.15in}
\end{figure*}

Although the above discussed \ac{dnn} variants have shown robustness against several anomalies, their defense capability is unexplored for data poisoning attacks. Existing methods often focus on detecting the poisoned data~\cite{bruckner2011stackelberg} or removing them through data sanitization~\cite{steinhardt2017certified}. These approaches have worked well, but they are computationally expensive and rely on the availability of a certified clean training dataset which may not always be available. To the best of our knowledge, the activation clustering (AC)~\cite{chen2018detecting} is the only other method that can clean a poisoned dataset without relying on a certified clean training dataset. This method relies on the assumption that a significant portion of the dataset is poisoned. However, such an assumption fails in realistic sparsely poisoned datasets (\textless10\%)~\cite{gu2017badnets}. We hypothesize that a deep-\ac{rbf} network trained on a sparsely poisoned dataset can be used for the discriminative ordering of clean and poisoned data without the need for a certified clean training dataset.

\section{Background}
\label{sec:background}

\subsection{Deep-\ac{rbf} network}
Deep-\ac{rbf} network is a conventional \ac{dnn} with an output layer of \ac{rbf} activation functions. A \ac{rbf} is a real-valued function that measures the distance of an input $x$ to some prototype vector. The similarity measure can be captured in the following definition of a \ac{rbf} unit using $\ell_p$-norm distance where $A\in\mathbb{R}^{n \times l}$, $b\in\mathbb{R}^l$, $x\in\mathbb{R}^n$ and $l \leq n$\cite{zadeh2018deep}.

\begin{equation}
\small
    \phi(x) = (||A^Tx+b||_p)^p
\end{equation}

In the context of a \ac{dnn}, \ac{rbf} units can be applied to the high-level features $f(x)$ extracted by the model from the raw input $x$ in order to classify the input into $k$ classes such that $k\in\{1,..,c\}$. Using the Euclidean metric and allowing $A=\mathbb{I}_n$, the deep-\ac{rbf} unit is defined as $\phi_k(x) = (||f(x)-W_k||_2)^2$


where $W_k\in\mathbb{R}^{|f(x)|}$ is a trainable weight vector intuitively representing the learned prototype of class $k$. The prediction category that results in the smallest distance is selected as the correct class during the evaluation phase. 


\subsubsection{\textbf{Training Loss Function}}
The deep-\ac{rbf} network can be trained using a metrics-learning inspired loss function named \textit{SoftML}, which is shown in \cref{eq:softml}. The function
was proposed in~\cite{zadeh2018deep} and is shown to avoid the vanishing gradient problem. 

\begin{equation}
\small
    J_{SoftML} = \sum_{i=1}^{N}{(\phi_{y_i}(x^{(i)}) + \sum_{j\not\in y_i}}log(1+e^{(\lambda - \phi_{y_i}(x^{(i)})})))
    \label{eq:softml}
\end{equation}

where $y_i$ is the correct class of input $x^{(i)}$, and $\lambda$ $>$ 0. The first term in the cost function aims at decreasing the distance between the prediction and the correct class. The second term aims at increasing the distance of the negative class. Further, as discussed in~\cite{zadeh2018deep}, the value of $\lambda$ has little effect on convergence and can be arbitrarily chosen.

\subsubsection{\textbf{Interpreting deep-\ac{rbf} network output}}
From a probabilistic point of view, the \cref{eq:softml} can be interpreted as the negative log-likelihood as discussed in~\cite{zadeh2018deep}. Therefore, the class prediction output of the deep-\ac{rbf} networks can be interpreted as non-normalized probabilities following the transformation.

\begin{equation}
\small
    P(\hat{y}=k|x) = \frac{e^{-\phi_k(x)}(1+e^{\lambda-\phi_k(x)})}{\prod_j(1+e^{\lambda-\phi_k(x)})}, \; k \in \{1,2,...,c\}
    \label{eq:probability}
\end{equation}

A rejection class $k=0$ can then be defined to capture the probability that $x$ belongs to no class in $\{1,2,...,c\}$.

\begin{equation}
\small
    P(\hat{y}=c+1|x) = \frac{1}{\prod_j(1+e^{\lambda-\phi_k(x)})}
    \label{eq:rejection}
\end{equation}



\subsection{Adversarial Examples}
The adversarial examples considered in this work are: (1) \textit{Physical attack}, which are adversaries that are physically realizable in the real world, and they include perturbing physical objects (e.g. traffic signs) in the images fed to the \acp{dnn}. In this work, we adapt the physical adversary introduced in~\cite{boloor2019simple}, where physical black lines are added at specific positions and angles to confuse an \ac{e2e} model to predict the wrong steering angles as shown in  \cref{fig:DeepNNCarAttack}; and, (2) \textit{Data poisoning attack}, which are adversaries that allow an attacker to modify the training procedure, alter the network's logic, or manipulate training dataset labels to encode a backdoor key. In this work, we adopt the \textit{injected pattern-key} attack where the training sample labels are altered whenever a backdoor key is encoded into the training input, allowing the attacker to exploit the attack by encoding the backdoor key into a test instance~\cite{chen2017targeted}.



\section{Deep-\ac{rbf} Network for Anomaly Detection}
\label{sec:framework}
In this section, we discuss the steps involved in detecting anomalies using deep-\ac{rbf} networks. The steps involved are: (1) problem transformation to transform a regression task to classification (this step can be skipped for classification tasks), (2) deep-\ac{rbf} design and training, and (3) anomaly detection. 



\subsection{Problem Transformation}
\label{sec:rbf-regression}
Certain \ac{cps} tasks (e.g. computing control actions) are regression based, and using the deep-\ac{rbf} network for anomaly detection requires transforming the regression task to classification. For explanation, we consider an example of a perception \ac{dnn} that predicts continuous steering predictions. The \ac{dnn} observes a sequence of images $\mathbb{X}_k=x_k \cdots x_{k-t}$ from the environment and predicts a continuous steering angle $s$ in range $[+\theta,-\theta]$. Here, $+\theta$ corresponds to a full right turn, and $-\theta$ corresponds to a complete left turn. This continuous steering angle needs to be discretized into n different classes $k$ = $\{0,1,....,n\}$. That is, each class corresponds to a small steering angle range ($s_k$) is calculated as $s_k = \frac{|+\theta|+|-\theta|}{n}$. 



where class k = 1 corresponds to a full right turn, and class k = n corresponds to a full left turn. The intermediate classes result in a right or left turn within $[+\theta,-\theta]$. The number of classes ($n$) for discretization is problem-specific, and it can impact the sensitivity of anomaly detection and control predictions, so it requires careful consideration. A Larger number of classes will provide fine-grained control over control predictions and results in high false negatives in detection. A smaller number of classes will make the detection insensitive resulting in high false positives.


\subsection{Deep-\ac{rbf} design and training}
As discussed earlier, a deep-\ac{rbf} network is a conventional \ac{dnn} with an \ac{rbf} layer attached to the output. The deep-\ac{rbf} network for the DeepNNCar regression task (continuous steering prediction) is shown in \cref{fig:DAVEIIArchitecture}. In this network, the convolutional layers extract the image features that are sent to an \ac{rbf} layer to perform classification. The number of \ac{rbf} units in the output layer corresponds to the number of classes ($n$) derived in the previous section (For a classification example, the number of \ac{rbf} units directly correspond to the number of classes in the task). The output of the deep-\ac{rbf} network has two components: (1) class predictions (e.g. steering class), and (2) rejection class probability that indicates if the input belongs to a known class or not.

The deep-\ac{rbf} network can then be trained using the \textit{SoftML} loss function (see \cref{eq:softml}). Training the network does not involve additional hyperparameters other than the standard ones like the number of epochs, batch size, learning rate, and the optimizer type. However, the high non-linearity of the \ac{rbf} units make it challenging to train the network. To address this, we apply the \ac{rbf} layer directly after the convolutional layers rather than after a series of fully connected layers. 

\subsection{Anomaly Detection}
During evaluation the operational test images are passed through the trained deep-\ac{rbf} network, and the output class predictions (discrete steering) is used to control the \ac{cps}, and the rejection class probability (\cref{eq:rejection}) is used along with a pre-selected threshold $\gamma$ to perform anomaly detection as shown in \cref{algo:algorithm1}. That is, if $P(\hat{y_t}=c+1|x_t) \geq \gamma$, the input can be identified to be an anomaly. 

\begin{algorithm}[!ht]
	\caption{Anomaly detection using deep-\ac{rbf} network}
    \textbf{Require}: deep-\ac{rbf} network $A_{RBF}$, Rejection threshold $\gamma$.\\
	\textbf{Input}: Image $x_t$ at time $t$. \\
	\textbf{Output}: $Anom_t$.
	\begin{algorithmic}[1]
	\STATE $P(\hat{y_t}=k|x_t), P(\hat{y_t}=c+1|x_t) = A_{RBF}(x_t)$
	\IF{$P(k=c+1|x) \geq \gamma$}
	        \STATE $Anom_t = 1$
	    \ELSE
	         \STATE $Anom_t = 0$
	\ENDIF
	\RETURN $Anom_t$
	\end{algorithmic} 
\normalsize{}
\label{algo:algorithm1}
\vspace{-0.05in}
\end{algorithm}









\section{Evaluation}
\label{sec:evaluation}
We evaluate \footnote{\label{fn:github}Jupyter notebooks to replicate the experiments can be found at https://github.com/Shreyasramakrishna90/RBF-Adversarial-Detection.git} the  performance of the deep-\ac{rbf} networks for detecting (a) black box physical attack on a hardware testbed called DeepNNCar that performs steering predictions (regression), and (b) data poisoning attack on the real-world \ac{gtsb} classification dataset.

\subsection{Detecting Physical Attack}
\label{sec:physicalattack}

\subsubsection{\textbf{Experimental Setup}}
In our first example, we implement the physical attack introduced in \cite{boloor2019attacking} on a hardware platform called DeepNNCar \cite{ramakrishna2019augmenting}. This testbed uses a Traxxas Slash 2WD 1/10 RC car and is computationally powered by a Raspberry Pi 3. The sensors on the car include a forward-looking camera and an IR opto-coupler attached to the rear wheel to measure the RPM and compute speed. The primary controller is NVIDIA's DAVE-II \ac{cnn} which uses the camera images to steer the car autonomously. The car is first manually driven to collect a training dataset that includes 6000 samples of images, steering PWM, and speed PWM values. The training dataset is split randomly into training, testing, and validation in a ratio of $70/15/15\%$. We then follow the steps in \cref{sec:rbf-regression} to transform this regression task into classification by discretizing the continuous steering labels into $10$ categories. Each discrete class represents a range of $6$\degree, allowing the car to turn discretely between $-30$\degree\space(sharp left, $y_i=0$) and $30$\degree\space(sharp right, $y_i=9$). 

\begin{figure}[t!]
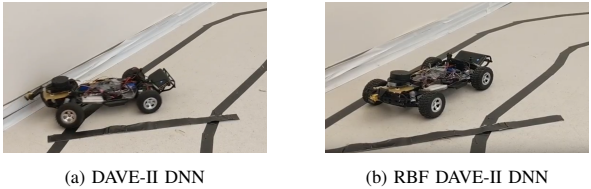

     \centering
     \begin{subfigure}[]{0.47\columnwidth}
         \includegraphics[width=3.5cm,height=2cm]{figures/physical_implementation.pdf}
         \centering
         \caption{DAVE-II \ac{dnn}}
         \label{fig:DAVEII_Deployed}
     \end{subfigure}
     \begin{subfigure}[]{0.47\columnwidth}
         \includegraphics[width=3.5cm,height=2cm]{figures/success_stop.pdf}
         \centering
         \caption{\ac{rbf} DAVE-II \ac{dnn}}
         \label{fig:RBF_Deployed}
     \end{subfigure}
     \caption{The physical attack caused DeepNNCar to crash when being controlled by the DAVE-II network; however, its \ac{rbf} extension was able to detect the anomaly and safely stop the car. Videos of these trial runs can be found at https://github.com/Shreyasramakrishna90/RBF-Adversarial-Detection.git} 
     
     \label{fig:DeepNNCarAttack}
     \vspace{-0.1in}
\end{figure}

We perform the physical attack by placing black lines across the track as shown in \cref{fig:DeepNNCarAttack}. The black lines are added at various angles to four distinct track sections like a left turn, a straight leading to a left turn, a right turn, and a straight leading to a right turn.

\subsubsection{\textbf{Competing Baselines}}
We compare two baseline networks to illustrate \acp{rbf} rejection class capability. The first is NVIDIA's DAVE-II regression network converted into a classification network ($k=10$) by adding $10$ fully connected neurons to the last layer followed by softmax activation. The other is the \ac{rbf} DAVE-II network that is designed by adding a hyperbolic tangent activation layer following the convolutional layers of the DAVE-II architecture and replacing the fully connected layers with an \ac{rbf} layer. 

We train these networks on $4200$ training images for $150$ epochs using adam optimizer, with categorical cross-entropy loss for the DAVE-II network and SoftML loss for \ac{rbf} DAVE-II network. For the \ac{rbf} DAVE-II network, we introduce a rejection threshold of $\gamma=0.6$, which is empirically selected to reduce false positives. 

\begin{figure}[t]
 \includegraphics[width=0.9\columnwidth]{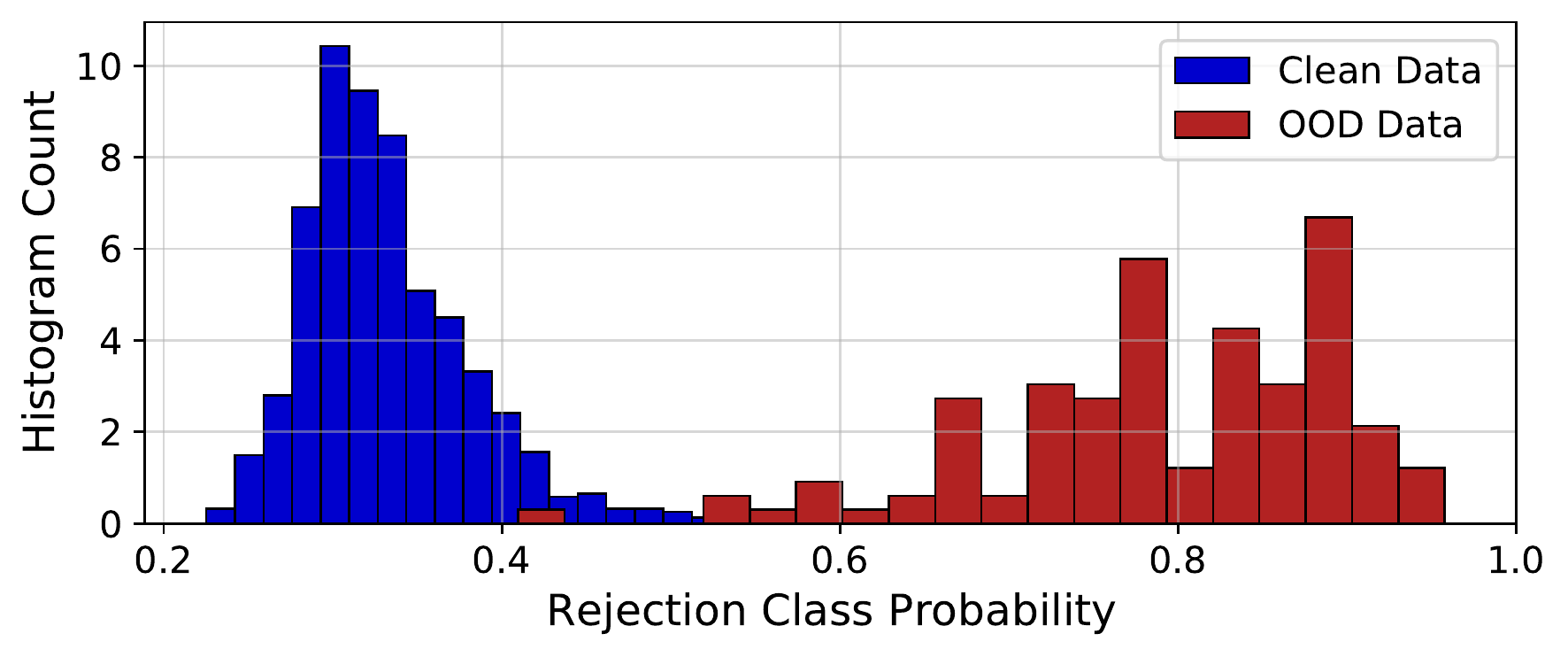}
 \centering
 \caption{A significant shift in the distribution of the rejection class probabilities was discovered for clean data and \ac{ood} (physical attack) data. We chose the threshold for the rejection class to be $\gamma = 0.6$.}
 \label{fig:DaveIIRejection}
 \vspace{-0.1in}
\end{figure}

\subsubsection{\textbf{Results}}

\begin{table}[t]
\centering
\renewcommand{\arraystretch}{1.2}
\footnotesize
\begin{tabular}{|c|c|c|}
\hline
\textbf{Actions}               & \textbf{\begin{tabular}[c]{@{}c@{}} RBF DAVE-II\end{tabular}} & \textbf{DAVE-II} \\ \hline
\textbf{Out-of-lane}           & 2                                                                    & 8                    \\ \hline
\textbf{Successful Navigation} & 3                                                                    & 4                    \\ \hline
\textbf{Safe Stop}             & 7                                                                    & NA                   \\ \hline
\end{tabular}
\caption{Performance of the DAVE-II and RBF DAVE-II in preventing the DeepNNCar from going out-of-lane because of the physical attack.}
\label{table:physicalattackresults}
\vspace{-0.1in}
\end{table}

We deploy the trained \ac{rbf} DAVE-II network on the DeepNNCar for runtime anomaly detection. We instruct the car to stop when the rejection class probability exceeded the rejection threshold. We also compared the performance of the two baseline networks in preventing the car from moving out-of-lane (see \cref{table:physicalattackresults}). We used each network to run the car for $12$ trial runs. For each trial, we approach the black attack lines at a constant speed of $0.5$ m/s and record the number of times the two networks lead the car out-of-lane.

We have classified the actions of the car into three classes to evaluate the performance of these networks. An \textit{Out-of-lane} is when the car moves out of the track lanes. A \textit{successful navigation} is when the car does not exit the track lanes but completes navigating the track. Finally, in the case of the \ac{rbf} DAVE-II network, a \textit{safe stop} is when the rejection class identifies a physical attack. In summary, the \ac{rbf} DAVE-II safely rejects the physical attack and executes the stop action as compared to the original DAVE-II network. Further, as shown in \cref{fig:DaveIIRejection}, the \ac{rbf} DAVE-II network exhibited a significant shift in the confidence of the rejection class for images with a physical attack.

\begin{table}[t]
\renewcommand{\arraystretch}{1.2}
\footnotesize
\centering
\begin{tabular}{|c|c|c|l|}
\hline
\textbf{Network}                                                              & \textbf{Precision (\%)} & \textbf{Recall (\%)} & \textbf{F1-score (\%)} \\ \hline
\textbf{RBF DAVE-II}                                                        & 96.4                    & 90.83                & 93.53                  \\ \hline
\textbf{\begin{tabular}[c]{@{}c@{}}VAE based\\ Reconstruction\end{tabular}} & 88.5                    & 90                   & 89.24                  \\ \hline
\end{tabular}
\caption{Performance comparison of the \ac{rbf} DAVE-II network and the \ac{vae} based reconstruction network for physical attack images.}
\label{table:comparison}
\vspace{-0.1in}
\end{table}

\textbf{(c) Comparison with other approaches}: We compare the \ac{rbf} DAVE-II with a reconstruction-based \ac{vae}~\cite{cai2020real} generative model. The \ac{vae} network has $5$ convolutional layers $24/36/48/48/64$ with 5 x 5 filters and 3 x 3 filters followed by one fully connected layer with $1164$ neurons. We use a symmetric deconvolutional decoder structure as the decoder. We trained the network for $150$ epochs using images from the training dataset. Finally, we evaluated the two networks on $1028$ images with physically attacked black lines that were collected from different trial runs. \cref{table:comparison} shows the performance of these networks in detecting the physical attack. As seen, the \ac{rbf} DAVE-II performs better in detecting the attack, and it also has smaller false positives in detecting images without attack.

\subsection{Detecting Data Poison Attack}

\begin{figure}[t]
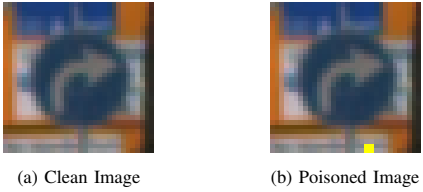

\centering
    \begin{subfigure}[b]{0.11\textwidth}
         \centering
         \includegraphics[width=\textwidth]{figures/clean_german_sign.pdf}
         \caption{Clean Image}
         \label{fig:clean_german}
     \end{subfigure}
     \hspace{1.3cm}
     \begin{subfigure}[b]{0.11\textwidth}
         \centering
         \includegraphics[width=\textwidth]{figures/poison_german_sign.pdf}
         \caption{Poisoned Image}
         \label{fig:poison_german}
     \end{subfigure}
     \caption{Examples of poisoned backdoor instances for the \ac{gtsb} dataset. The poisoned image has a yellow post-it like note.}
     \vspace{-0.1in}
\end{figure}

\begin{figure*}[t]
 \includegraphics[width=0.85\textwidth]{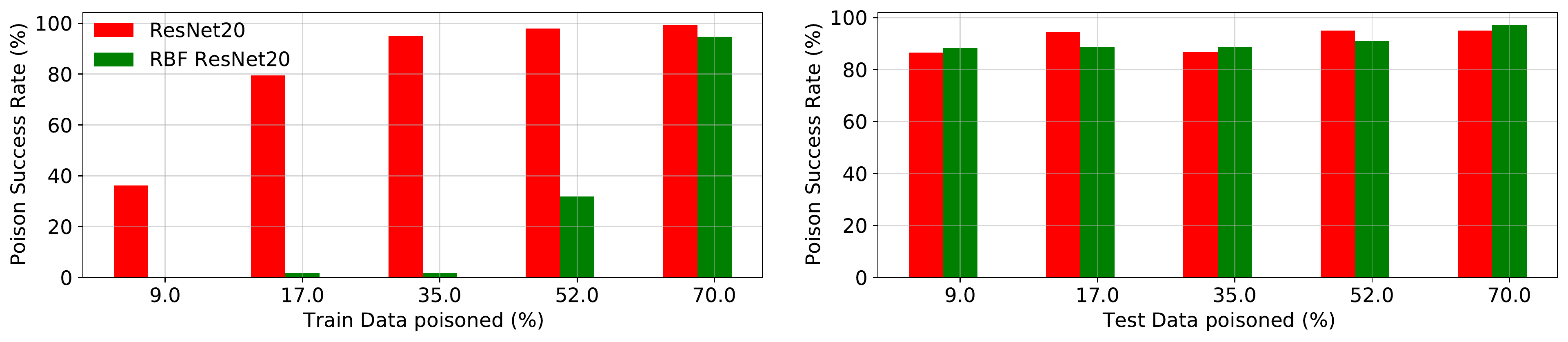}
 \centering
 \caption{The poison attack success rate for ResNet20 and \ac{rbf} ResNet20 networks. (left) train data poisoned using $1200$ backdoor key instances, and (right) test data poisoned using the same $1200$ instances. The \ac{rbf} ResNet20 network only starts to fail only when $>35\%$ of the training data gets poisoned.}
 \label{fig:poison_results}
 \vspace{-0.1in}
\end{figure*}

\subsubsection{\textbf{Experimental Setup}}
We performed the poisoning attack on the \ac{gtsb} dataset \cite{GermanTrafficSignDataSet} that has over $50,000$ images of traffic signs from $43$ traffic sign classes. We adapt the popular \textit{injected pattern-key} attack where the attack is a targeted label attack that attempts to cause a \ac{dnn} to predict any road sign as an $80$ km/hr signboard (see \cref{fig:poison_german}), whenever a backdoor key similar to a post-it note is encoded in the image. We poison the dataset by (1) randomly select $n_p$ instances outside of the $80$ km/h image data, (2) add a post-it like note at a random location in the image, and (3) change the instance label to that of the $80$ km/h road sign. A poisoning attack is successful if the model predicts non-poisoned images as their ground truth and poisoned images as the modified label.

\subsubsection{\textbf{Competing Baselines}}
We compare two baseline networks to illustrate \acp{rbf} rejection class capability. First is the ResNet20 \cite{he2016deep} network which has $20$ convolutional layers and one layer of softmax activation functions. The other is the \ac{rbf} ResNet20 network, which replaces the final fully-connected layer with an \ac{rbf} layer preceded by a hyperbolic tangent function. We split the \ac{gtsb} dataset into a training dataset of $39209$ images and an evaluation dataset with $11430$ clean images and $1200$ backdoor instances. We poisoned the training dataset while the evaluation dataset is kept clean. We then trained the networks on the training dataset for $150$ epochs using categorical cross-entropy loss for the ResNet20 network and SoftML loss for the \ac{rbf} ResNet20 network.

\subsubsection{\textbf{Results}}

\textbf{(a) Robustness evaluation:} We evaluate the robustness of the baseline networks by adjusting the number of poisoned samples ($n_p$) in the training dataset. We incrementally increase the poisoned images in the training data and record the poison attack success rates for both networks. As seen in \cref{fig:poison_results}, the poisoning success rate of ResNet20 is greater than $30\%$ after only $5\%$ of the train data has been poisoned, and the success rate increases to $90\%$ after $35\%$ of the data is poisoned. In comparison, \ac{rbf} ResNet20 requires a larger amount of the data to be poisoned for the attack to be successful. The success rate is negligible after $35\%$ of the train data is poisoned, and it increases to $30\%$ only after $53\%$ of the train data is poisoned. However, both the networks succumb to the attack after $70\%$ of the data is poisoned. Further, in \cref{fig:poison_results}-b, we find that both ResNet20 and \ac{rbf} ResNet20 achieve similar overall accuracy on the test data. This finding eliminates the possible argument that ResNet20 is simply learning the data distribution better than \ac{rbf} ResNet20 and is, therefore, more likely to be successfully poisoned.


\begin{figure*}[t]
 \includegraphics[width=0.9\textwidth]{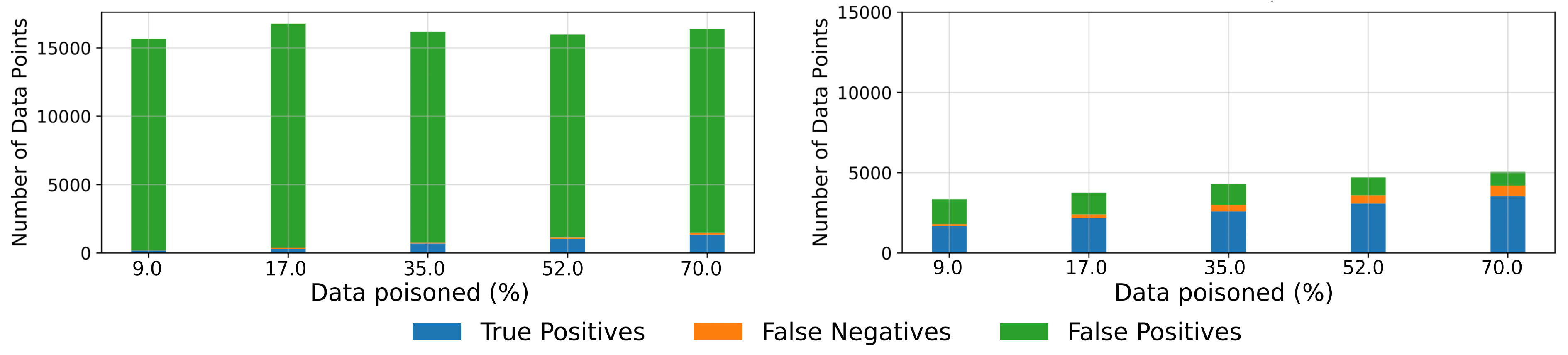}
 \centering
 \caption{(left) AC method using K-means ($n_c=2$) and PCA ($|D|=10$), (right) \ac{rbf} ResNet20 network's rejection capability with $\beta=1.72$. The plots show the capability of the two methods in correctly classifying the training samples as \textit{poisoned} or \textit{clean}.}
 \label{fig:poison_cleaning}
 \vspace{-0.1in}
\end{figure*}

\textbf{(c) Comparison with other approaches:} We compare the \ac{rbf} ResNet20 network to the activation clustering (AC) method \cite{chen2018detecting} which clusters the penultimate layer's activations to separate poisoned and clean instances. To perform the AC method, we use the author's suggestion of K-means ($k=2$) and PCA to reduce the penultimate layer's activations to $10$ dimensions.  For the \ac{rbf} ResNet20 network, a rejection threshold of $\beta=1.72$ was used to modestly cover the tail end of the distribution of $\phi_{y_i}(X^i_{poison})$. 

\cref{fig:poison_cleaning} shows the results of adjusting $n_p$ and comparing the two cleaning methods for the poisoned \ac{gtsb} dataset. In the sparsely poisoned conditions, the \ac{rbf} ResNet20 network was able to achieve on average higher true positive rates and lower false-positive rates than the AC method. Even at lower values of $n_p$ where the poisoning success rate on the regular classifiers still exceeds $90\%$, the AC method tends to predict fewer true positives and a significant number of false positives exceeding $15000$. However, the \ac{rbf} ResNet20 network (see \cref{fig:poison_cleaning} (right)) has a higher true positive and lower false positives (not exceeding $5000$) for different values of $n_p$. The results show that the \ac{rbf} ResNet20 network is highly robust to sparsely poisoned data and begins to slowly fail as the value of $n_p$ increases, whereas the AC method dramatically fails. 

\subsection{Resource Evaluation}
We performed the resource evaluations on a desktop with AMD Ryzen Threadripper 16-Core Processor, $4$ NVIDIA Titan Xp GPU, and $128$ GiB memory. For these evaluations, we compare two different approaches. The first is an \ac{rbf} DAVE-II network that performs both discrete steering predictions and anomaly detection. The second is the original NVIDIA DAVE-II regression network for continuous steering predictions and a reconstruction based \ac{vae} for anomaly detection. The structure of the \ac{vae} is discussed in \cref{sec:physicalattack}.

\subsubsection{\textbf{Execution Time}}
The DAVE-II network took an average of $65.4$ milliseconds for steering angle predictions, and the reconstruction-based \ac{vae} network took an average of $53$ milliseconds for anomaly detection. In comparison, the \ac{rbf} DAVE-II network only took an average of $44$ milliseconds for both discrete steering predictions and anomaly detection. In summary, the \ac{rbf} DAVE-II network has a $62.8\%$ reduction in the execution time compared to the \ac{vae}. This reduction is because of the reduced operations the \ac{rbf} DAVE-II network has to perform following the convolutional layers. 

\subsubsection{\textbf{Memory Usage}}
The DAVE-II network utilized an average memory of $2.0$ GB, and the reconstruction-based \ac{vae} network used an average memory of $3.6$ GB. In comparison, the \ac{rbf} DAVE-II network only utilized an average memory of $1.1$ GB. The \ac{rbf} DAVE-II network used less memory because of the fewer operations following the convolutional layers, compared to the \ac{vae} network that has a bulky encoder-decoder architecture that requires higher computations.

\section{Conclusion and Future Work}
\label{sec:conclusion}
This paper evaluates the efficiency of deep-\ac{rbf} networks for detecting \ac{dnn} related anomalies in \ac{cps} applications. We propose the use of a single deep-\ac{rbf} network to perform both controller predictions and anomaly detection in \ac{cps} regression tasks. However, the use of \ac{rbf} functions limits the network's applicability only to classification tasks. So, we discuss the steps in converting a \ac{cps} regression task (continuous steering prediction) to a classification task (discrete steering prediction) and then train a deep-\ac{rbf} network for class prediction and anomaly detection. To support our hypothesis, we evaluated the deep-\ac{rbf} networks for two different attacks on \ac{cps} regression and classification tasks. Our results have shown the deep-\ac{rbf} network to robustly detect these attacks with minimal resource requirement. 

Future extensions of the deep-\ac{rbf} networks include: (1) extending the rejection capability to different types of \ac{ood} data (e.g. brightness, occlusion, etc.), and (2) using the rejection capability for high level controller selection.

\textbf{Acknowledgement}: This work was supported by the DARPA Assured Autonomy project and Air Force Research Laboratory. 

\balance
\bibliographystyle{IEEEtran}
\bibliography{main.bib}

\begin{thebibliography}{10}
\providecommand{\url}[1]{#1}
\csname url@samestyle\endcsname
\providecommand{\newblock}{\relax}
\providecommand{\bibinfo}[2]{#2}
\providecommand{\BIBentrySTDinterwordspacing}{\spaceskip=0pt\relax}
\providecommand{\BIBentryALTinterwordstretchfactor}{4}
\providecommand{\BIBentryALTinterwordspacing}{\spaceskip=\fontdimen2\font plus
\BIBentryALTinterwordstretchfactor\fontdimen3\font minus
  \fontdimen4\font\relax}
\providecommand{\BIBforeignlanguage}[2]{{%
\expandafter\ifx\csname l@#1\endcsname\relax
\typeout{** WARNING: IEEEtran.bst: No hyphenation pattern has been}%
\typeout{** loaded for the language `#1'. Using the pattern for}%
\typeout{** the default language instead.}%
\else
\language=\csname l@#1\endcsname
\fi
#2}}
\providecommand{\BIBdecl}{\relax}
\BIBdecl

\bibitem{bojarski2016end}
M.~Bojarski, D.~Del~Testa, D.~Dworakowski, B.~Firner, B.~Flepp, P.~Goyal, L.~D.
  Jackel, M.~Monfort, U.~Muller, J.~Zhang \emph{et~al.}, ``End to end learning
  for self-driving cars,'' \emph{arXiv preprint arXiv:1604.07316}, 2016.

\bibitem{cai2020real}
F.~Cai and X.~Koutsoukos, ``Real-time out-of-distribution detection in
  learning-enabled cyber-physical systems,'' in \emph{2020 ACM/IEEE 11th
  International Conference on Cyber-Physical Systems (ICCPS)}.\hskip 1em plus
  0.5em minus 0.4em\relax IEEE, 2020, pp. 174--183.

\bibitem{sundar2020out}
V.~K. Sundar, S.~Ramakrishna, Z.~Rahiminasab, A.~Easwaran, and A.~Dubey,
  ``Out-of-distribution detection in multi-label datasets using latent space of
  $\beta$-vae,'' \emph{arXiv preprint arXiv:2003.08740}, 2020.

\bibitem{hartsell2021resonate}
C.~Hartsell, S.~Ramakrishna, A.~Dubey, D.~Stojcsics, N.~Mahadevan, and
  G.~Karsai, ``Resonate: A runtime risk assessment framework for autonomous
  systems,'' \emph{arXiv preprint arXiv:2102.09419}, 2021.

\bibitem{amini2018variational}
A.~Amini, W.~Schwarting, G.~Rosman, B.~Araki, S.~Karaman, and D.~Rus,
  ``Variational autoencoder for end-to-end control of autonomous driving with
  novelty detection and training de-biasing,'' in \emph{2018 IEEE/RSJ
  International Conference on Intelligent Robots and Systems (IROS)}.\hskip 1em
  plus 0.5em minus 0.4em\relax IEEE, 2018, pp. 568--575.

\bibitem{goodfellow2014explaining}
I.~J. Goodfellow, J.~Shlens, and C.~Szegedy, ``Explaining and harnessing
  adversarial examples,'' \emph{arXiv preprint arXiv:1412.6572}, 2014.

\bibitem{zadeh2018deep}
P.~H. Zadeh, R.~Hosseini, and S.~Sra, ``Deep-rbf networks revisited: Robust
  classification with rejection,'' \emph{arXiv preprint arXiv:1812.03190},
  2018.

\bibitem{crecchi2020fader}
F.~Crecchi, M.~Melis, A.~Sotgiu, D.~Bacciu, and B.~Biggio, ``Fader: Fast
  adversarial example rejection,'' \emph{arXiv preprint arXiv:2010.09119},
  2020.

\bibitem{ramakrishna2019augmenting}
S.~Ramakrishna, A.~Dubey, M.~P. Burruss, C.~Hartsell, N.~Mahadevan,
  S.~Nannapaneni, A.~Laszka, and G.~Karsai, ``Augmenting learning components
  for safety in resource constrained autonomous robots,'' in \emph{2019 IEEE
  22nd International Symposium on Real-Time Distributed Computing
  (ISORC)}.\hskip 1em plus 0.5em minus 0.4em\relax IEEE, 2019, pp. 108--117.

\bibitem{bruckner2011stackelberg}
M.~Br{\"u}ckner and T.~Scheffer, ``Stackelberg games for adversarial prediction
  problems,'' in \emph{Proceedings of the 17th ACM SIGKDD international
  conference on Knowledge discovery and data mining}, 2011, pp. 547--555.

\bibitem{steinhardt2017certified}
J.~Steinhardt, P.~W. Koh, and P.~Liang, ``Certified defenses for data poisoning
  attacks,'' \emph{arXiv preprint arXiv:1706.03691}, 2017.

\bibitem{chen2018detecting}
B.~Chen, W.~Carvalho, N.~Baracaldo, H.~Ludwig, B.~Edwards, T.~Lee, I.~Molloy,
  and B.~Srivastava, ``Detecting backdoor attacks on deep neural networks by
  activation clustering,'' \emph{arXiv preprint arXiv:1811.03728}, 2018.

\bibitem{gu2017badnets}
T.~Gu, B.~Dolan-Gavitt, and S.~Garg, ``Badnets: Identifying vulnerabilities in
  the machine learning model supply chain,'' \emph{arXiv preprint
  arXiv:1708.06733}, 2017.

\bibitem{boloor2019simple}
A.~Boloor, X.~He, C.~Gill, Y.~Vorobeychik, and X.~Zhang, ``Simple physical
  adversarial examples against end-to-end autonomous driving models,''
  \emph{arXiv preprint arXiv:1903.05157}, 2019.

\bibitem{chen2017targeted}
X.~Chen, C.~Liu, B.~Li, K.~Lu, and D.~Song, ``Targeted backdoor attacks on deep
  learning systems using data poisoning,'' \emph{arXiv preprint
  arXiv:1712.05526}, 2017.

\bibitem{boloor2019attacking}
A.~Boloor, K.~Garimella, X.~He, C.~Gill, Y.~Vorobeychik, and X.~Zhang,
  ``Attacking vision-based perception in end-to-end autonomous driving
  models,'' \emph{arXiv preprint arXiv:1910.01907}, 2019.

\bibitem{GermanTrafficSignDataSet}
J.~Stallkamp, M.~Schlipsing, J.~Salmen, and C.~Igel, ``The {G}erman {T}raffic
  {S}ign {R}ecognition {B}enchmark: A multi-class classification competition,''
  in \emph{IEEE International Joint Conference on Neural Networks}, 2011, pp.
  1453--1460.

\bibitem{he2016deep}
K.~He, X.~Zhang, S.~Ren, and J.~Sun, ``Deep residual learning for image
  recognition,'' in \emph{Proceedings of the IEEE conference on computer vision
  and pattern recognition}, 2016, pp. 770--778.

\end{thebibliography}
\end{document}